\pdfoutput=1
\documentclass{article}
\usepackage[margin=0.7in]{geometry}
\usepackage[parfill]{parskip}
\usepackage[utf8]{inputenc}
\usepackage{authblk}

\usepackage{amsmath,amssymb,amsfonts,amsthm}

\usepackage{graphicx}
\usepackage{multirow}
\usepackage{booktabs}

\usepackage{subfig}

\title{Advancing Text-to-GLOSS Neural Translation Using a Novel Hyper-parameter Optimization Technique}
\date{}

\author{Ouargani Younes\thanks{ younes\_ouargani@um5.ac.ma}\qquad El Khattabi Noussaima\thanks{ e.noussaima@um5r.ac.ma}}
\affil{Laboratory of Conception and Systems (Electronics, Signals, and Informatics), Faculty of Sciences Rabat, Mohammed V University Rabat. }
\begin{document}
\maketitle
\begin{abstract}

    In this paper, we investigate the use of transformers for Neural Machine Translation of text-to-GLOSS for Deaf and Hard-of-Hearing communication.
    Due to the scarcity of available data and limited resources for text-to-GLOSS translation,
    we treat the problem as a low-resource language task.
    We use our novel hyper-parameter exploration technique to explore a variety of architectural parameters and build an optimal transformer-based architecture specifically tailored for text-to-GLOSS translation.
    The study aims to improve the accuracy and fluency of Neural Machine Translation generated GLOSS.
    This is achieved by examining various architectural parameters including layer count, attention heads, embedding dimension, dropout, and label smoothing to identify the optimal architecture for improving text-to-GLOSS translation performance.
    The experiments conducted on the PHOENIX14T dataset reveal that the optimal transformer architecture outperforms previous work on the same dataset.
    The best model reaches a ROUGE (Recall-Oriented Understudy for Gisting Evaluation) score of 55.18\% and a BLEU-1 (BiLingual Evaluation Understudy 1) score of 63.6\%, outperforming state-of-the-art results on the BLEU1 and ROUGE score by 8.42 and 0.63 respectively.

\end{abstract}

\textit{\textbf{KEYWORDS}}\par
{\textit{Sign Language, Auditory impairment, Text to GLOSS, neural machine translation}}

\section{introduction}

Speech to Sign Language translation plays a vital role in facilitating effective communication in the Deaf and Hard of Hearing (DHH) community.
It allows speech to be transformed into a visual representation of sign language, facilitating communication and inclusion for DHH individuals.
However, speech-to-sign language translation is a challenging task due to several factors, such as the complexity of sign languages. This is because sign languages use not only manual signals (i.e. hand gestures) but also grammatical and semantic features such as facial expressions, mouth morphemes, eye gaze, and body movements.
This makes it extremely difficult to successfully translate speech into sign language.
The need for multiple intermediate translations also contributes to the difficulty of this task, as cumulative translation errors lead to a high translation error rate, resulting in unusable translations.
The lack of datasets and scarcity of resources in this area also contributes to hindering progress in the translation task.
To address this issue, recent research has explored different approaches to neural machine translation of text-to-GLOSS.
Advanced deep learning techniques have been used such as Gated Recurrent Units (GRU) \cite{stoll_sign_2018,stoll_text2sign_2020,amin_sign_2021}, Long Short Term Memory (LSTM) \cite{amin_sign_2021}, Generative Adversarial Networks (GAN) \cite{stoll_sign_2018,stoll_text2sign_2020}, and Transformers \cite{saunders_progressive_2020}.
By carefully tuning the parameters of each architecture, researchers improve translation performance and ensure that the resulting GLOSS output accurately captures the linguistic nuances and specific requirements for a correct translation.

In this paper, we contribute to the fast-growing body of research by investigating the use of Transformers for text-to-GLOSS translation.
Our research focuses on leveraging the strength of the Transformer architecture in NMT to address the text-to-GLOSS translation as a low-resource translation task.
We explore multiple parameters to design a specifically tailored Transformer-based architecture for this task, and use the PHOENIX14T \cite{forster_extensions_2014} dataset to evaluate the performance of different architectural choices and identify the most prominent parameters that contribute to improving the translation accuracy, and identify the most effective configuration for achieving accurate and fluent GLOSS outputs.

In this context, the primary contributions of this paper can be outlined as follows:
\begin{itemize}
    \item We propose a novel approach for hyper-parameter exploration on low-resource tasks, and use it to create a Transformer-based text-to-GLOSS sequence-to-sequence translation model.
    \item We evaluate the performance of the proposed model on the phoenix-2014T corpus using multiple metrics, such as BLEU (BiLingual Evaluation Understudy), and ROUGE (Recall-Oriented Understudy for Gisting Evaluation).
    \item We show that (to the best of our knowledge) our architecture outperforms state-of-the-art models on the PHOENIX-2014T corpus.
\end{itemize}

\section{Methodology}

\subsection{Tranformers}

In this section, we delve into the architecture of our transformer-based model for text-to-GLOSS translation,
drawing inspiration from the ground-breaking encoder-decoder structure common in neural sequence transduction models \cite{cho_learning_2014-1,bahdanau_neural_2016-3}.
This approach is crucial for maintaining the sequential nature of the task, allowing the generation of coherent and contextually accurate GLOSS outputs.

Interwoven with an intricate attention mechanism, an encoder, and a decoder are the two fundamental components of our model architecture.
A sequence of input symbol representations \( (x_1,..., x_n )\) is mapped by the encoder into an output sequence  \( (z_1,..., z_n )\),
whereas the decoder generates the output sequence \( (y_1,..., y_n )\) autoregressively.
This is done to preserve the temporal dependencies inherent in the translation process, hence the model generates each output element based on the previously generated symbols.

\begin{figure}[h]
    \centering
    \includegraphics[width=0.45\textwidth]{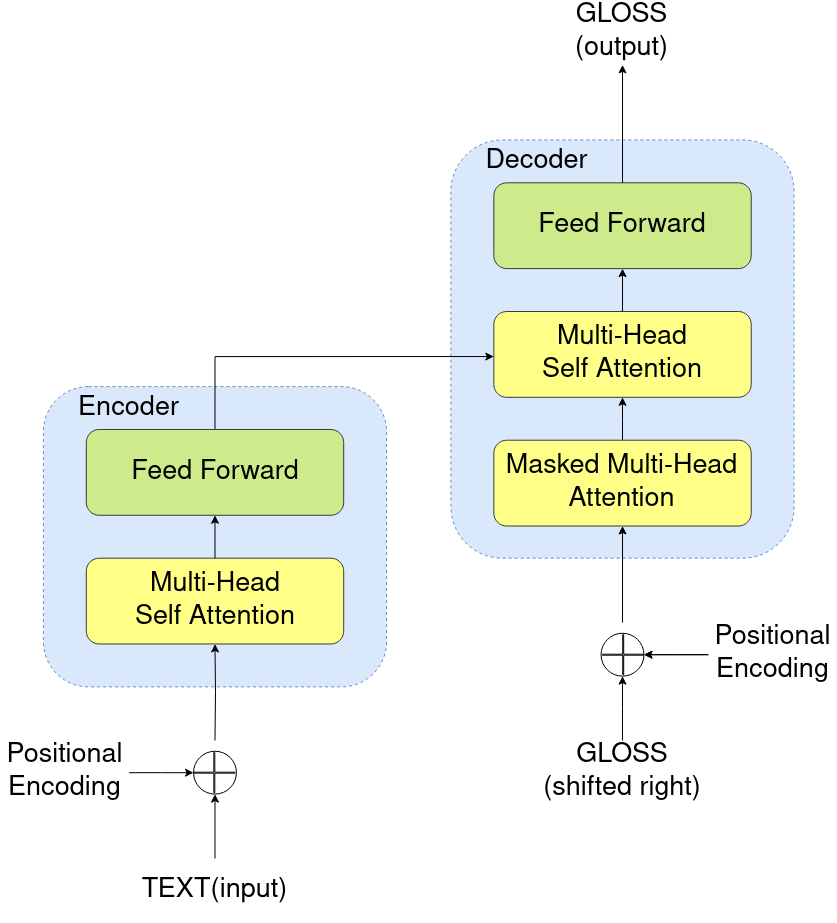}
    \caption{Proposed encoder-decoder model}
    \label{fig_trans}
\end{figure}

The Transformer architecture, renowned for its prowess in handling sequential data through stacked self-attention and point-wise, fully connected layers is represented in Figure~\ref{fig_trans} with both the encoder and the decoder half represented.

The encoder-decoder architecture is structured as follows:
Our encoder comprises a stack of \textit{N} identical layers, each housing two sub-layers.
The first sub-layer features a multi-head self-attention mechanism, while the second harbors a simple position-wise fully connected feed-forward network.
A residual connection \cite{he_deep_2016} is applied around each sub-layer, followed by a layer normalization\cite{ba_layer_2016}, ensuring smooth information flow.
Notably, all sub-layers in the model produce an output of dimension \textit{d}(embedding dimension), matching the embedding layer dimension.
In a similar fashion to the encoder, the decoder is constructed as a stack of \textit{N} identical layers, with the addition of a sub-layer that handles multi-head attention over the encoder stack's output.
The residual connection and layer normalization are retained, hence the output of each sub-layer in both the encoder and the decoder can be formulated as: \[LayerNorm(x)=(x+Sublayer(x))\] with \textit{LayerNorm} the layer normalization operation, and \textit{Sublayer} the function applied by the sublayer.
In order to maintain the auto-regressive property of the decoder, the self-attention sub-layer is masked to prevent information flow from subsequent positions.
Offsetting the output embeddings by one position, and masking guarantees that predictions for position \textit{i} only depend on known outputs preceding position \textit{i}.

\subsection{Hyper-parameter exploration}

In this section, we propose a novel transformer-based architecture specifically tailored for text-to-GLOSS translation, taking into consideration the difficulties presented by the resource-constrained nature of the task.
Due to their proficiency in modeling long-range dependencies and successfully capturing contextual information, transformers have shown exceptional effectiveness in a variety of natural language processing tasks, particularly in neural machine translation (NMT).
But despite their massive success in NMT, their potential in the text-to-GLOSS translation task has not been fully realized.
However,  this can't be achieved without hyper-parameter optimization, as it is a crucial step to achieve good performance in low-resource tasks.

A complete hyper-parameter exploration to find the optimal Transformer architecture can be prohibitively expensive using grid search, thus researchers often resort to either a random hyper-parameter search  \cite{bergstra_random_2012} or a grid search for one hyper-parameter at a time.
Despite yielding better hyper-parameter combinations, random search tends to require a lot more time to reach a complete value range exploration.
The grid search for one hyper-parameter at a time on the other hand, for a hyper-parameter x, only finds the optimal value for the current hyper-parameter set, and this value remains unchanged despite changing other parameters.
This led us to use a consecutive hyper-parameter exploration, as a parameter could be optimal using the current parameter set but could lead to a sub-optimal architecture upon changing one or more parameters.
This is done using the hyper-parameter range in Table~\ref{tab_param_space}.

\begin{table}[h]
    \centering
    \begin{tabular}{ll}
        \toprule
        Hyper-parameter        & Values \\
        \midrule
        number of layers& 1 2 3 4 5 6 7     \\
        feed-forward dimension & 128    256     512 \\
        embedding dimension    &   32   64  128\\
        attention heads      &   1  2   4   8 \\
        dropout &   0.1 0.2 0.3 0.4 0.5\\
        batch-size &  256   512 1024    2048    4096\\
        label smoothing &    0.1    0.2 0.3 0.4 0.5 0.6\\
        warmup steps &  100 200 300 400 500 600\\
        \bottomrule
    \end{tabular}
    \caption{Explored hyper-parameter space}
    \label{tab_param_space}
\end{table}

In the process of consecutive hyper-parameter exploration, we adopt a systematic approach to fine-tune the transformer-based architecture. 
This process is performed by carefully selecting an initial set of hyper-parameters, then each parameter is optimized sequentially.
In this step, we focus on one hyper-parameter at a time while keeping others fixed, and assess the performance of the model on the chosen dataset.
The hyper-parameter values are adjusted based on the results obtained from the first round of optimization.
The finetuning process is iterative, this procedure is repeated for each hyper-parameter, one by one, until no further improvement of the model output is observed.
By employing this consecutive hyper-parameter exploration method, we achieve a better understanding of the contribution of each hyper-parameter to the model's performance and develop an optimal architecture that maximizes translation accuracy by using the most effective hyper-parameter set.

\subsection{Dataset}

For running our experiments, we selected the PHOENIX14T \cite{forster_extensions_2014}  parallel text-GLOSS corpus for the task of text-to-GLOSS translation in sign language.
It is part of the RWTH-PHOENIX-Weather 2014 corpus, which was developed by the Human Language Technology \& Pattern Recognition Group from RWTH Aachen University in Germany.

\begin{table}[]
    \centering
    \begin{tabular}{ccccccc}
        \toprule
        \multirow{2}{*}{} & \multicolumn{3}{c}{GLOSS} & \multicolumn{3}{c}{TEXT} \\
        \cmidrule(lr){2-4}\cmidrule(lr){5-7}
        & Train   & Dev    & Test   & Train   & Dev    & Test  \\
        \midrule
        sentences         & 7 096   & 519    & 642    & 7 096   & 519    & 642   \\
        words             & 67 781  & 3 745  & 4 257  & 99 081  & 6 820  & 7 816 \\
        vocabulary        & 1 066   & 393    & 411    & 2 887   & 951    & 1 001\\
        \bottomrule
    \end{tabular}
    \caption{Statistics of the PHOENIX14T dataset}
    \label{tab_phoe}
\end{table}

The corpus comprises high-quality video recordings of German sign language interpretation extracted from the daily news and weather forecasts from the years 2009-2011.
Moreover, speech recognition coupled with manual cleaning has been used to transcribe the original German speech.
Furthermore, manual GLOSS notation for weather forecasts from 386 editions is provided in German Sign Language (DGS).
More details of the corpus are provided in table \ref{tab_phoe}.

The PHOENIX14T dataset was chosen for several compelling reasons:
\begin{itemize}

    \item Non-synthetic: The dataset contains real-world sign language interpretations performed by professional interpreters.
        This is essential to train and evaluate our model in a realistic setting, using data from actual scenarios encountered in sign language communication.

    \item Data quality: The dataset contains high-quality data signed and manually transcribed by experts, this guarantees accurate data for optimizing our architecture. The high quality of the recordings ensures accuracy and reliability in capturing the sign language transcriptions, allowing our model to learn from reliable and precise examples.

    \item Widely used benchmark: The PHOENIX14T dataset is a well-known and widely used benchmark for sign language translation in the DHH research community.
        The adoption of this dataset allows us to compare our results with other state-of-the-art models and establish a standardized evaluation metric for our proposed architecture.

\end{itemize}

The PHOENIX14T dataset was selected for our work for its realism, high quality, and it's wide use as a GLOSS-text translation benchmark.
Leveraging this non-synthetic parallel text-glos corpus enables us to provide valuable contributions to the field of sign language translation, and enhance communication and accessibility for the DHH community.

\section{Experimental results}

In this section, we present the results obtained from our experiments.
First, we describe the evaluation metrics used to evaluate our work, then we outline the hyper-parameter exploration procedure and the proposed model, and finally, we report the experimental results obtained using the PHOENIX14T dataset, and compare them to prior studies.

\subsection{Evaluation metrics}

Two evaluation metrics are used to score our work; BLEU (BiLingual Evaluation Understudy) and ROUGE (Recall-Oriented Understudy for Gisting Evaluation).
The BLEU and ROUGE scores are commonly used in Natural Language Processing tasks, such as machine translation and text summarization.

BLEU is a popular metric used to evaluate machine translation quality by comparing the translation outputs to the reference text.
It compares the overlap between the reference text and the translation in terms of n-grams (consecutive sequences of words).
BLEU computes precision scores for n-grams of varying lengths (up to 4-grams) and then combines the scores to an overall BLEU score while taking into consideration a brevity penalty to prevent overfitting of sentence lengths.
A higher BLEU score indicates a better translation quality as the predicted output aligns well with the reference translation.
ROUGE on the other hand is a set of metrics mainly used for text summarization, but it's often used to report machine translation results.
ROUGE evaluates the quality of generated translations or summaries by comparing the predicted text to the reference summaries or translations.
Similar to BLEU, ROUGE measures the overlap of n-grams. In addition to that, it also measures the recall scores assessing the important information coverage in the predicted output.
The ROUGE and BLEU scores serve as a widely recognized benchmark in the NLP community,
making it easier to assess the performance of our models and compare our results with existing state-of-the-art research.

\begin{figure}[h]
    \centering
    \includegraphics[width=0.75\textwidth]{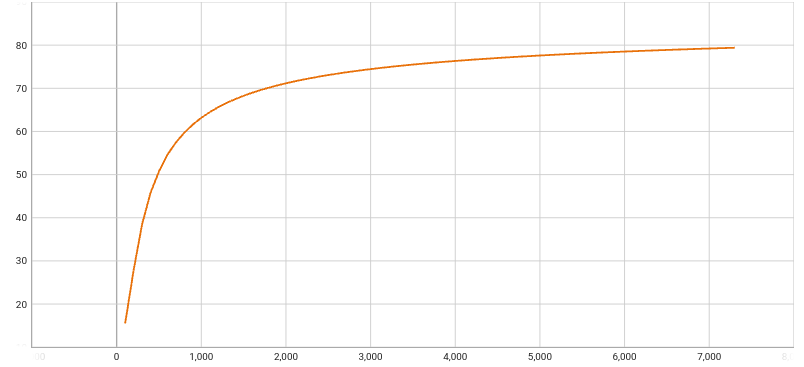}
    \caption{Model accuracy on Train set}
    \label{fig_train_acc}
\end{figure}

\begin{figure}[h]
    \centering
    \includegraphics[width=0.75\textwidth]{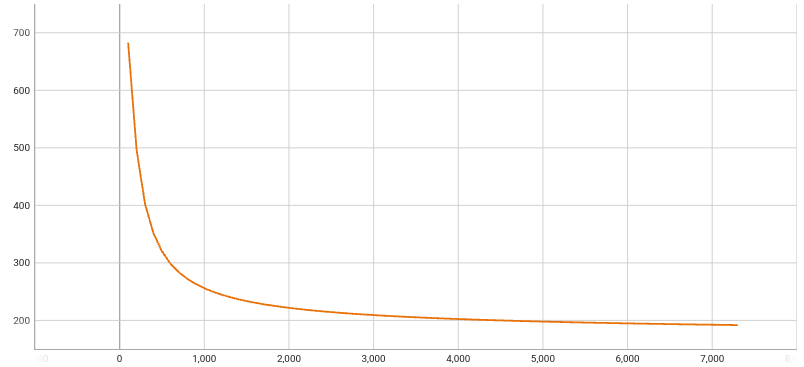}
    \caption{Model perplexity on Train set}
    \label{fig_train_ppl}
\end{figure}

\begin{figure}[h]
    \centering
    \includegraphics[width=0.75\textwidth]{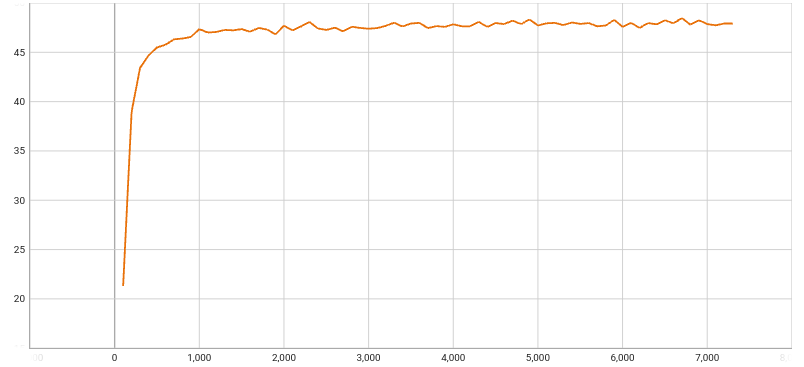}
    \caption{Model accuracy on Test set}
    \label{fig_test_acc}
\end{figure}

\begin{figure}[h]
    \centering
    \includegraphics[width=0.75\textwidth]{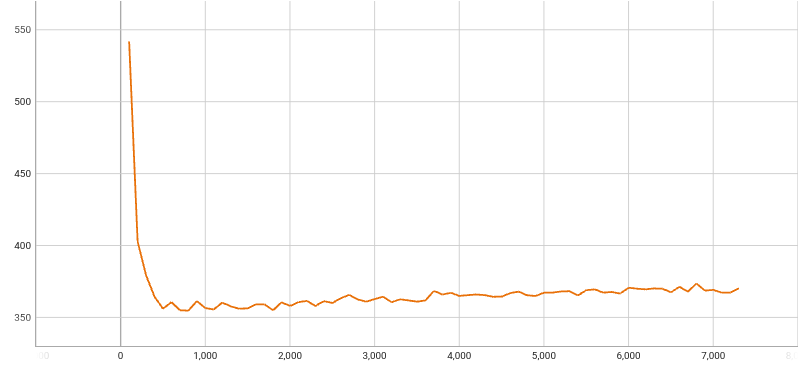}
    \caption{Model perplexity on Test set}
    \label{fig_test_ppl}
\end{figure}

\subsection{Results and discussion}

As described earlier, our consecutive hyper-parameter exploration starts by choosing an initial transformer architecture (the initial hyper-parameter set can be chosen randomly or manually generated by leveraging previously acquired expertise).
Sequential parameter optimization is then manually conducted by focusing on individual parameters at a time while keeping the remaining parameters fixed.
For instance, we start by finetuning the feed-forward dimension, isolating its impact on the model's performance.
Based on the results of the initial round of optimization, we retain the most optimal hyper-parameter values.
The iterative exploration process, through multiple consecutive runs, ensures a comprehensive coverage of the hyper-parameter space as displayed in Table~\ref{tab_param_space}.
This process is continued for each hyper-parameter until we reach an optimal architecture.

For training the model, adam \cite{kingma_adam_2017} optimizer was used with a learning rate of 0.001, 0.9 for beta1, and 0.998 for beta2.
The learning rate decay method was set to Noam \cite{vaswani_attention_2017}, with 300 warmup steps.
Inevitably, signs of overfitting appeared early in the experimentation phase, to resolve this problem, multiple hyper-parameters had to be introduced to the parameter set.
The initial parameter set contained: feed-forward dimension, embedding dimension, attention head number, batch size, and warmup steps.
In addition to these hyper-parameters, dropout, and label smoothing were introduced to the hyper-parameter set to ensure a better-performing model with reduced overfitting.
Both dropout and label smoothing seem to have a positive effect on the training as they significantly improve the ROUGE and BLEU scores.
The best dropout value in our experiment is 0.3, while label smoothing provides more score improvement at a value of 0.6.

\begin{table}[h]
    \centering
    \begin{tabular}{ll}
        \toprule
        Hyper-parameter        & Value \\

        \midrule
        number of layers&   5    \\
        feed-forward dimension & 256      \\
        embedding dimension    &  64     \\
        attention heads      &    2 \\
        dropout &      0.3 \\
        batch-size &    4096   \\
        label smoothing &   0.6    \\
        warmup steps &  300 \\
        \bottomrule
    \end{tabular}
    \caption{Optimal Transformer architecture}
    \label{tab_best_arch}
\end{table}

From our experiments, we observed that deeper transformers coupled with smaller feed-forward layer dimensions resulted in a better-performing model compared to shallower transformers with bigger feed-forward layer dimensions.
In \cite{raganato_fixed_2020} authors show that fixing the number of attention heads to 1 during the training doesn't cause any performance degradation for mid-sized datasets, this lines up with our observations as more attention heads produced a lower ROUGE score compared to models with less attention heads.

Table\ref{tab_best_arch} shows our best-performing model, and it reached a ROUGE score of 55.18\% and a BLEU 1 score of 63.6\%.

However, as observed in Figures~\ref{fig_train_acc} and \ref{fig_test_acc} despite having the best performance, the best model (i.e. the model with the best ROUGE score) only manages to reach 77.21\% training accuracy, which translates to 47.35\% test accuracy.
The test accuracy is much lower than the training accuracy despite the use of regularization techniques, we think this value disparity between the train and test accuracy is due to the low-resource dataset as the limited resources make generalization harder for the trained model.
However, we can observe from Figures~\ref{fig_train_ppl} and \ref{fig_test_ppl} that perplexity on the test set is minimal around the 1,050 step where the model reaches optimal BLEU 1 performance, however, the best ROUGE score was obtained on step 4900.

The performance of our model is compared to prior studies in Table~\ref{tab_sota_comp},
all the results in the table are obtained from the test set of the PHOENIX14T text-GLOSS parallel corpus.
In addition to the optimal ROUGE score, we also provide the optimal BLEU score for different n-grams from 1 to 4 to provide a better insight into the translation performance.
Furthermore, each system's architecture is also provided.
The best BLEU score obtained is 19.04\%.
When it comes to ROUGE scores, our system outperforms both the attention-equipped Recurrent Neural Network-based architecture proposed in \cite{stoll_text2sign_2020}, and the Gated Recurrent Unit with attention proposed in \cite{amin_sign_2021}, with a ROUGE score increase of 7.08, and 12.22 respectively.
Our system also marginally outperforms the Symbolic Transformer proposed in \cite{saunders_progressive_2020}, providing a ROUGE score improvement of 0.63.
As for BLEU scores, our system provides a higher BLEU1 score than all the other methods, with a BLEU1 score difference of as much as 19.7 compared to \cite{amin_sign_2021}.
The BLEU1 score difference between our model and the proposed models in \cite{stoll_text2sign_2020}, and \cite{saunders_progressive_2020} is 12.93, and 8.42 respectively.
In the BLEU2, BLEU3, and BLEU4 scores, our model offers comparable performance to the model in \cite{amin_sign_2021} as the scores were 28.5, 15.2, and 9.0 for the BLEU2, BLEU3, and BLEU4 respectively.

\begin{table}[h]
    \centering
    \begin{tabular}{  p{25em}  c  cccc}
        \toprule
        \multirow{2}{*}{Methods} & \multirow{2}{*}{ROUGE} & \multicolumn{4}{c}{BLEU}\\
        \cmidrule(lr){3-6}
        & & BLEU1 & BLEU2 & BLEU3 & BLEU4 \\
        \midrule
        Recurrent Neural Network with Luong attention\cite{stoll_text2sign_2020} & 48.10 & 50.67 & 32.25 & 21.54 & 15.25 \\
        Gated Recurrent Unit with Luong attention\cite{amin_sign_2021} & 42.96 & 43.90 & 26.33 & 16.16 & 10.42 \\
        Transformer\cite{saunders_progressive_2020} & 54.55 & 55.18 & \textbf{37.10} & \textbf{26.24} & \textbf{19.10} \\
        Our proposed approach & \textbf{55.18} & \textbf{63.6} & 28.5 & 15.2 & 9.0 \\
        \bottomrule
    \end{tabular}
    \caption{ROUGE and BLEU scores of PHOENIX14T corpus test set}
    \label{tab_sota_comp}
\end{table}

\section{Conclusion}

Deaf and Hard of Hearing people rely on sign language as their primary means of communication.
But despite recent advances in Neural Machine Translation, speech/text-to-sign language translation is still an active area of research.
In this work, we proposed a novel approach for developing a text-to-sign language GLOSS translation model.
Our approach is based on consecutive hyper-parameter exploration and fine-tuning to develop a machine-learning architecture specifically tailored for this task.

Our research compared the influence of the explored hyper-parameters on the translation quality, we used a transformer-based architecture and explored 8 different hyper-parameters: number of layers, feed-forward dimension, embedding dimension, number of attention heads, dropout, batch size, label smoothing, warmup steps.
For our experiments, we used the PHOENIX14T text to German Sign Language (DGS) parallel corpus, and both the ROUGE and BLEU scores to benchmark our models.
Our best-performing models reached a ROUGE score of 55.18\% and a BLEU 1 score of 63.6\%.
The proposed method yields a high-accuracy model outperforming state-of-the-art models in both ROUGE and BLEU-1 scores.

The proposed system can be used in different fields such as education, health, and public transit.
It can be especially useful as a lot of research focuses on sign language recognition and sign-to-text translation.
For future work, more hyper-parameters can be explored such as BPE, optimizers, and attention dropout.

\bibliographystyle{unsrt}
\bibliography{MyLibrary}

\end{document}